\documentclass[letterpaper, 10 pt, conference]{ieeeconf}
\IEEEoverridecommandlockouts
\let\labelindent\relax
\usepackage{enumitem}
\usepackage{balance}
\usepackage[font={small}]{caption}
\usepackage{subcaption}
\usepackage{array}
\usepackage{textcomp}
\usepackage{mathtools, nccmath}
\usepackage{graphicx}
\usepackage{amsfonts}
\usepackage{amsmath}
\usepackage{amssymb}
\usepackage{algorithm}
\usepackage{algorithmic}
\usepackage{hyperref}
\usepackage{tikz}
\usepackage{arydshln}
\usepackage{multirow}
\usepackage{bm}
\usepackage{epstopdf}
\usepackage{cite}
\usepackage{siunitx}

\newcommand{\update}[1]{{\color{black} #1}}

\title{\LARGE \bf Robotic Guide Dog: Leading a Human with Leash-Guided \\ Hybrid Physical Interaction}
\author{ Anxing Xiao*, Wenzhe Tong*, Lizhi Yang*, Jun Zeng, Zhongyu Li, and Koushil Sreenath
\thanks{* Authors have contributed equally.}
\thanks{All authors are with the Department of Mechanical Engineering, University of California, Berkeley, CA, 94720, USA, 
\tt\small\{xax, wenzhe.t99, lzyang, zengjunsjtu, zhongyu\_li, koushils\}@berkeley.edu}
}

\begin{document}
\maketitle
\begin{abstract}
An autonomous robot that is able to physically guide humans through narrow and cluttered spaces could be a big boon to the visually-impaired. 
Most prior robotic guiding systems are based on wheeled platforms with large bases with actuated rigid guiding canes.
The large bases and the actuated arms limit these prior approaches from operating in narrow and cluttered environments. 
We propose a method that introduces a quadrupedal robot with a leash to enable the robot-guiding-human system to change its intrinsic dimension (by letting the leash go slack) in order to fit into narrow spaces. 
We propose a hybrid physical Human Robot Interaction model that involves leash tension to describe the dynamical relationship in the robot-guiding-human system. 
This hybrid model is utilized in a mixed-integer programming problem to develop a reactive planner that is able to utilize slack-taut switching to guide a blind-folded person to safely travel in a confined space. 
The proposed leash-guided robot framework is deployed on a Mini Cheetah quadrupedal robot and validated in experiments (Video\footnote{ Video: \urlstyle{same}\url{https://youtu.be/FySXRzmji8Y}}). 
\end{abstract}

%% INTRODUCTION
\section{Introduction}
\begin{figure}
    \centering
    \includegraphics[width=0.8\linewidth]{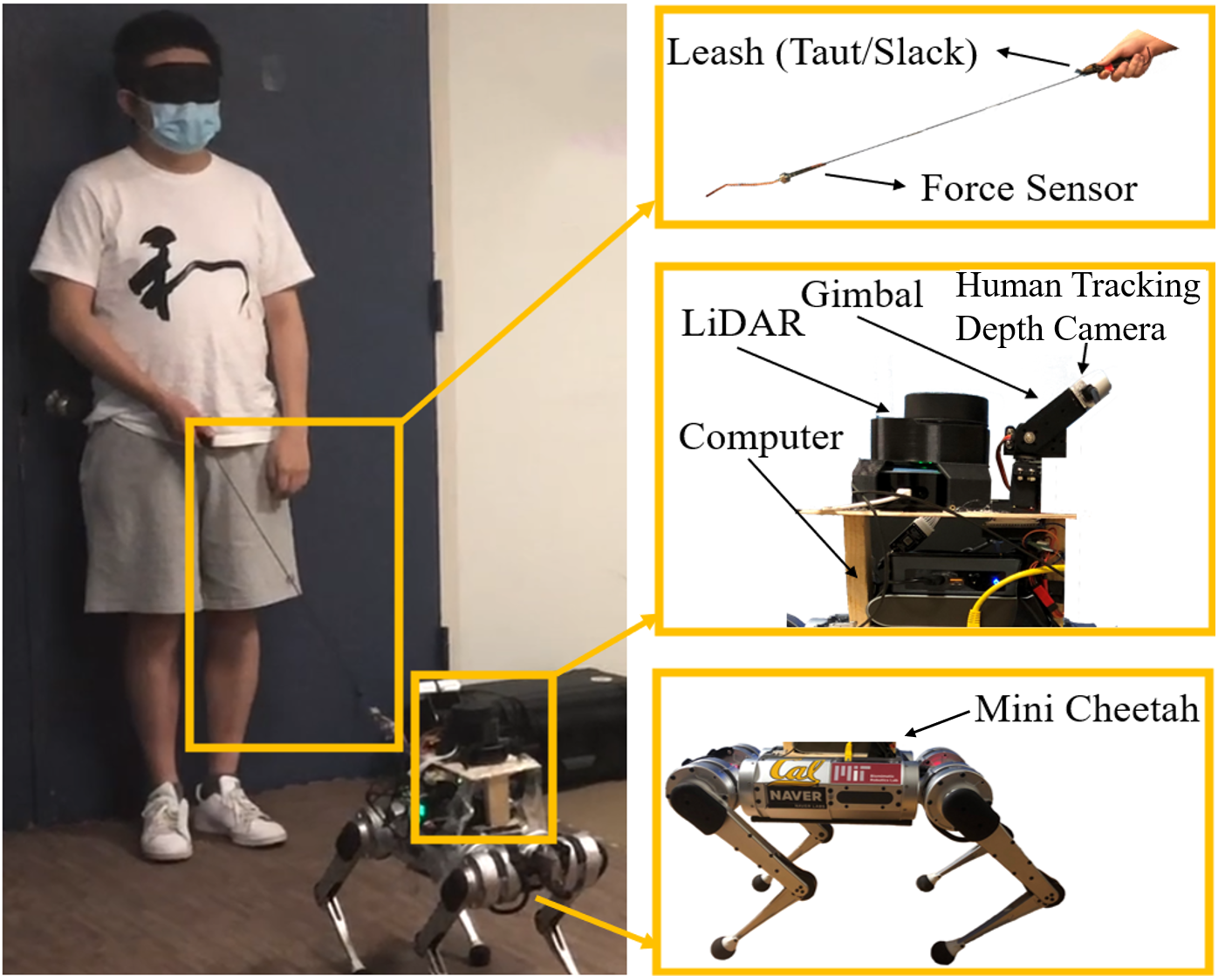}
    \caption{The Mini Cheetah is guiding a blindfolded person to avoid obstacles with leash-guided assistance: a leash (top right) is used to connect between the robot and the human, a 2D LiDAR is used for robot localization and a depth camera is used for human detection (middle right). The leash could be taut or slack during the navigation.
}
    \label{fig:cover}
    \vspace{-0.5cm}
\end{figure}

Guide dogs play a critical role in our society by helping the frail, elderly, or visually impaired people navigate the world. % is meaningful to the minority group in the society.
However, a well-behaved guide dog usually needs to be selected and trained individually.
In addition, the skills from one dog cannot be transferred to another one.
This makes training guide dogs both time and labor intensive with the process not easily scalable.
With recent progress in robotics, an autonomous robot could potentially take over this responsibility. 
Our goal in this paper is to create a robotic guide dog. 
Most previous guide robots have large foot-bases \cite{morris2003robotic, Palopoli_2015Dalu,wachaja2017navigating} and usually require an actuated rigid arm to guide the human \cite{borenstein1997guidecane, chuang2018deep, zhongyu_2019}, which results in limited capabilities of operating in narrow spaces.
Moreover, the usage of a rigid arm brings an additional layer of complexity in mechanical and control design.
A small robot that could guide humans with a leash could potentially solve such an issue.
The ability of the leash to become slack allows the robot to change the internal dimensions of the human-robot system, and thus allows the robot to guide the human through narrow spaces, such as a doorway.
%%ZL:11.1.2020 say something why the a robot with leash can address this issue. the dimension can change? what scenario will be useful for leash addressed
However, utilizing a leash could involve a hybrid system switch, i.e., the leash could be taut or slack, which makes this motion planning more challenging.
Therefore, we seek to address such a problem where we utilize a quadrupedal robot, a Mini Cheetah \cite{katz2019mini}, to guide a visually-impaired person via a leash to navigate in narrow spaces, as shown in Fig. \ref{fig:cover}.

% Recently, a human-centered approach \cite{zhongyu_2019} allows a ballbot to physically to guide people, where the interaction between the robot and the human is through a rigid rod.
% The rigid rod was also widely used in other work and provides a well-established geometric relation between human and the robot, but it diminishes its capabilities to guide human to navigate through narrow, crooked environments.
% Moreover, the ballbot robot is only able to function in a flat ground.
% To handle these challenges, a more direct way is to introduce leash connection between them, but the leash could become taut or slack and it becomes very challenging to solve the motion planning problem as now the human-robot system becomes hybrid \cite{foehn2017fast, zeng2020differential}. In this paper, we handle this hybrid system by the intuition that human will stay immobile when there is no tension in the leash, and move along the direction of leash when the tension is reestablished. Additionally, a quadrupedal robot is applied in our robot guide system, which was proven to interact with different environment in previous work \cite{bosworth2016robot, katz2019mini, yang2020dynamic}. All together above leads us to provide an end-to-end hybrid quadruped robot guiding system which could lead human to navigate in narrow clustered and rugged environments.

\subsection{Related Work}
\subsubsection{Robotic Guide Dog}
Using robots to guide humans is a long-studied problem, and the mainstream approaches employ either a robotic cane  \cite{borenstein1997guidecane,chuang2018deep,Ye_2016Robotic}, or a robotic walker \cite{morris2003robotic, Palopoli_2015Dalu,wachaja2017navigating}.
However, for \cite{borenstein1997guidecane,chuang2018deep,Ye_2016Robotic}, since they use an actuated rigid arm as a guiding cane between the robot and the human, the system may get stuck in a region which cannot fit the robot arm.
Moreover, an actuated arm brings more DoFs and needs additional effort on control and mechanical design.
%% ZL:11.1.2020 why flexibility is reduced? Need more control and design efforts? addressed
Of the approaches with robot walkers, \cite{morris2003robotic, Palopoli_2015Dalu} are designed for the elderly and movement-impaired population and do not consider the visually impaired, and while \cite{wachaja2017navigating} considers the visually impaired, the guiding system has a very large foot-base. The large base also occurs in \cite{morris2003robotic, Palopoli_2015Dalu}, and makes maneuvering and guiding a person in a narrow space impossible.
Apart from using the movement of a rigid robot arm, different ways to interact between the robot and the human have
%% ZL:11.1.2020 why human-robot interaction is mentioned here? You mean during leading? say something that why you mention this, and why your method can outperform those mentioned method. addressed
also been explored, with most methods employing either vocal cues~\cite{Yang2020IndoorQS}, or haptic feedback~\cite{Katzschmann_2018}.
Some~\cite{dakopoulos_2009_wearable, zhongyu_2019, chuang2018deep} take it a step further and employ a mix of the methods mentioned above.
However, the vocal cues may not always be clear to the people being led, and as previously mentioned, rigid rods decrease the mobility of the system in a confined space significantly. While haptic feedback has been explored in \cite{Katzschmann_2018}, it is only a wearable system with some vibration feedback, and does not consider a system containing both a human and a movable robot. Thus, the prior work does not guide a human while also being able to change the intrinsic dimension of the human-robot system. 
%All the previous work doesn't provide a direct guidance to the human while being able to change the intrinsic dimension of the robot.
% Thus in order to solve this problem, we present a leash-guided method to lead the human, and the intuition follows that if the leash is taut, then the people being led will follow along the direction of the leash tension, and when the leash is slack, the user does not move. This is more optimal than using a rigid rod since it's flexible and can go into environments where the rigid rod is ineffective.
%% ZL:11.1.2020 too long here and not fit to the introduction, move to the methodology part
%% ZL:11.1.2020 whats your hri method? physical hri? so what will be the advantage of phri? I will suggest to say somethig good about haptics. Previous work speech, vibration, whats the disadvantage? addressed
\subsubsection{Hybrid System Planning}
Hybrid system control and planning is challenging for physical human-robot interaction (pHRI) tasks \cite{pervez2008safe}.
There is some prior work on hybrid system \emph{control} in pHRI \cite{del2012online, magrini2016hybrid}. 
For path \emph{planning} in pHRI, it was demonstrated in \cite{ulrich2000vfh, kulyukin2006robot, palopoli2015navigation, zhongyu_2019} that a collision-free trajectory could be generated to guide the human. 
As we introduce a leash for the robot to guide the human, the system becomes hybrid as the leash could be taut or slack.
For hybrid modes on leash tension, previous works about aerial systems formulate the path planning either through a special mechanical design \cite{zeng2019geometric}, mixed-integer programming \cite{tang2015mixed} or collocation-based optimization with complementarity constraints \cite{foehn2017fast, zeng2020differential}.
However, physical human-robot interaction is not considered in \cite{tang2015mixed, magrini2016hybrid, foehn2017fast, zeng2020differential}, and hybrid path planning for pHRI for applications using  mobile robots still remains an open problem. 

\subsection{Contributions}
We make the following contributions:
\begin{itemize}
    \item One of the first end-to-end hybrid physical human-robot interaction (hybrid-pHRI) framework is presented for a robotic guide dog  with a leash.
    \item A hybrid model is developed to capture the dynamic relationship in the robot-leash-human system, involving a leash tension model. The hybrid model is validated with experimental data.
    %Both these two models are validated with experimental data.\li{what 2 models?}
    \item We formulate a mixed-integer programming problem in the path planner to safely guide humans to avoid obstacles during navigation to the goal location while also considering the taut/slack modes of the leash.
    \item We validate our hybrid-pHRI robot guide framework experimentally on a quadrupedal robot, where a Mini Cheetah is empowered to navigate with a collision-free trajectory in narrow environments while guiding people by exploiting hybrid mode switches.
\end{itemize}

\section{Physical Human Robot Interaction (pHRI) Model}
\label{sec:hybrid-pHRI-model}
The ability of the robot to determine the current configuration of the human-robot system and to interact with the human via the leash is very important for successful path planning and guiding a human. Thus a pHRI model is needed to capture the state of the human-robot system and serve as the underlying basis for planning and interaction.
%% ZL:11.1.2020 say something about why you need a model and this section? addressed
\subsection{Human-robot System}
\begin{figure}[t]
    \centering
    \includegraphics[width=0.8\linewidth]{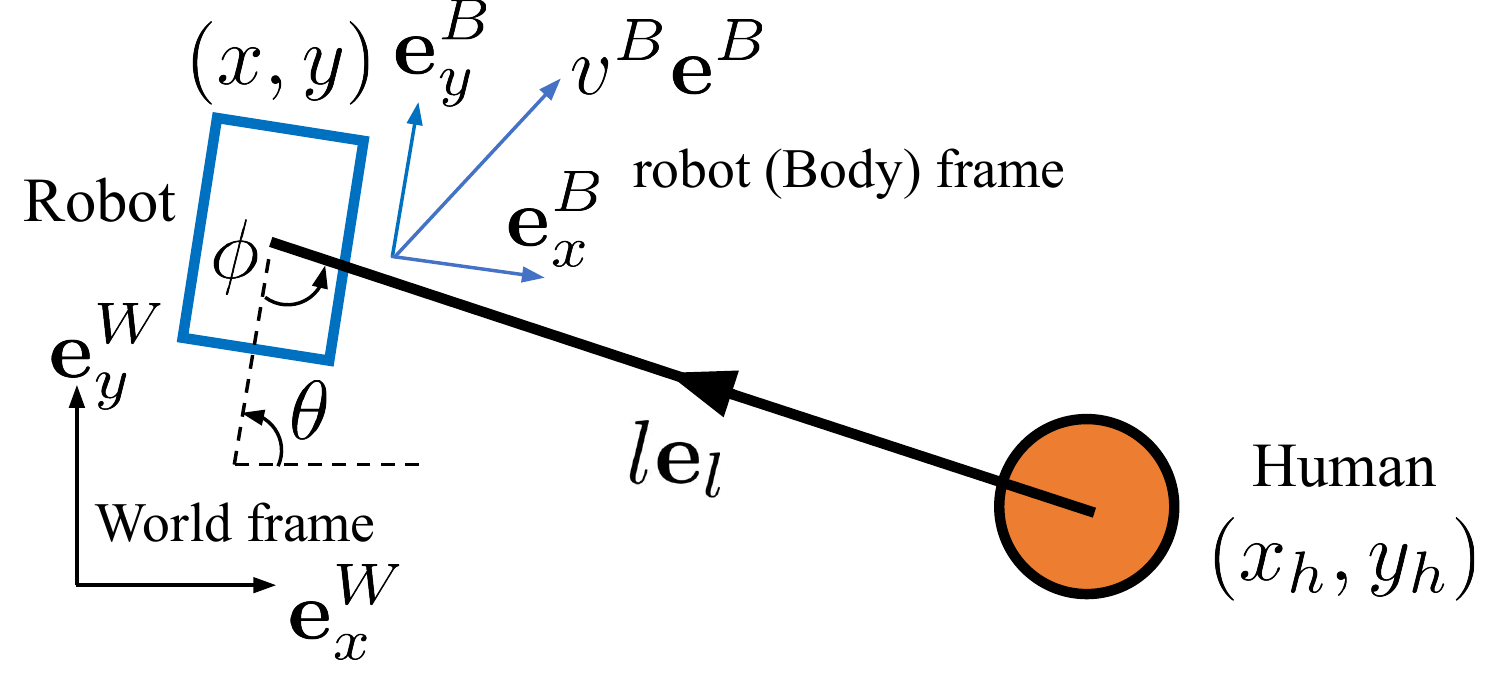}
    \caption{Configuration of the human-robot guiding system. Human $(x^h, y^h)$ is guided by a leash attached to the robot $(x, y)$, $l\vec e_l$ represents the relative position between them.}
    % \zli{need to be more informative, say something like robot is leading a human by a leash(what is leash in the figure)}
    \label{fig:configuration}
    \vspace{-0.3cm}
\end{figure}

The human-robot system configuration is defined as follows:
\begin{equation}
    \label{eq:human-robot-geometric-relation}
    \mathbf{x}^h = \mathbf{x} - l \mathbf{e}_l,
\end{equation}
where $\mathbf{x}^h = (x^h, y^h)$ and $\mathbf{x} = (x, y)$ are the position of the human and robot respectively. Furthermore, $l$ represents the distance between the human and the robot and $\mathbf{e}_l = (\cos(\theta - \phi), \sin(\theta - \phi)) \in S^1$ is the unit vector point from the human to the robot along the leash. $\theta$ represents the orientation of the robot in the world frame $W$, and $\phi$ represents the relative orientation of the human in the robot body frame $B$.
This configuration is valid no matter if the leash is taut or slack, as shown in Fig. \ref{fig:configuration}.
When the leash is taut, the system has four degrees-of-freedom with configuration space $Q_t = \mathbb{R}^2 \times S^{1}$ and $l = l_0$.
When the leash is slack, the system has five degrees-of-freedom with configuration space $Q_s = \mathbb{R}^2 \times S^{1} \times \mathbb{R}$ and $l$ becomes less than $l_0$. Here $l_0$ is the length of the leash.

\subsection{Hybrid Dynamic Model}
\label{subsec:hybrid-dynamics}

\subsubsection{Taut Mode}
The assumption for our hybrid dynamic model comes from our intuitive hypothesis that the human will be guided by the robot and move in the direction of the force applied by the leash when the leash is taut.
Moreover, when the leash is slack, the human will not feel any force in the leash and thus will not move. Based on this assumption, we define our hybrid system with continuous-time generalized coordinates $\mathbf{q} = (x, y, \theta, \phi, l) \in \mathbb{R}^5$ and with an input of $\mathbf{u}= (\mathbf{v}^B, \omega) \in \mathbb{R}^3$, where $\mathbf{v}^B = v^B \mathbf{e}^B$ represents the robot velocity in its body frame, shown in Fig. \ref{fig:configuration}. The robot velocity $\mathbf{v}^B$ and angular velocity $\omega$ are the commands which could be sent to the Mini Cheetah directly and a tracking controller could track these commands.

When the leash becomes taut, based on our assumption and the geometric relationship, the hybrid dynamics $\dot{\mathbf{q}}_t = f_{t}(\mathbf{q}_t,\mathbf{u}_t)$ could be formulated as follows,
\begin{subequations}
\label{eq:leash-taut-dynamics}
\begin{align}
     \dot{x} &= \alpha_x v^B \mathbf{e}^B \cdot \mathbf{e}^W_x,\\
     \dot{y} &= \alpha_y v^B \mathbf{e}^B \cdot \mathbf{e}^W_y,\\
     \dot{\theta} &= \alpha_{\theta} \omega,\\
     \dot{\phi} &= -\alpha_{\theta} \omega - \alpha_{\phi} \| v^B \mathbf{e}_l \times \update{\mathbf{e}^B}  \|/l_0, \label{eq:leash-taut-phi-dynamics}\\
     \update{l} &= l_{0},
\end{align}
\end{subequations}
shown in the left mode in Fig. \ref{fig:hybridsystem}. 

The human-robot interaction and inelastic collision when the leash switches from slack to taut is very hard to model and this external force brings disturbance for the robot tracking controller. To compensate these tracking errors from this disturbance, we introduce 
\begin{equation}
\label{eq:alpha-vector}
\bm{\alpha} = \left[\alpha_x, \alpha_y, \alpha_{\theta}, \alpha_{\phi} \right] \in \left[0, 1\right]^4
\end{equation}
as the discount coefficients in \eqref{eq:leash-slack-dynamics}. These discount coefficients are smaller than one since human always tend to drag the leash from the opposite direction with respect to the commands to the robot, i.e.,  $\update{v^B} \mathbf{e}^B$ and $\update{\omega}$. These discount coefficients allow us to capture the unknown interaction acting on the robot from the human with a four-dimension representation and these four coefficients which can be tuned for a good prediction.

\subsubsection{Slack Mode}
\begin{figure}
    \centering
    \includegraphics[width=0.9\linewidth]{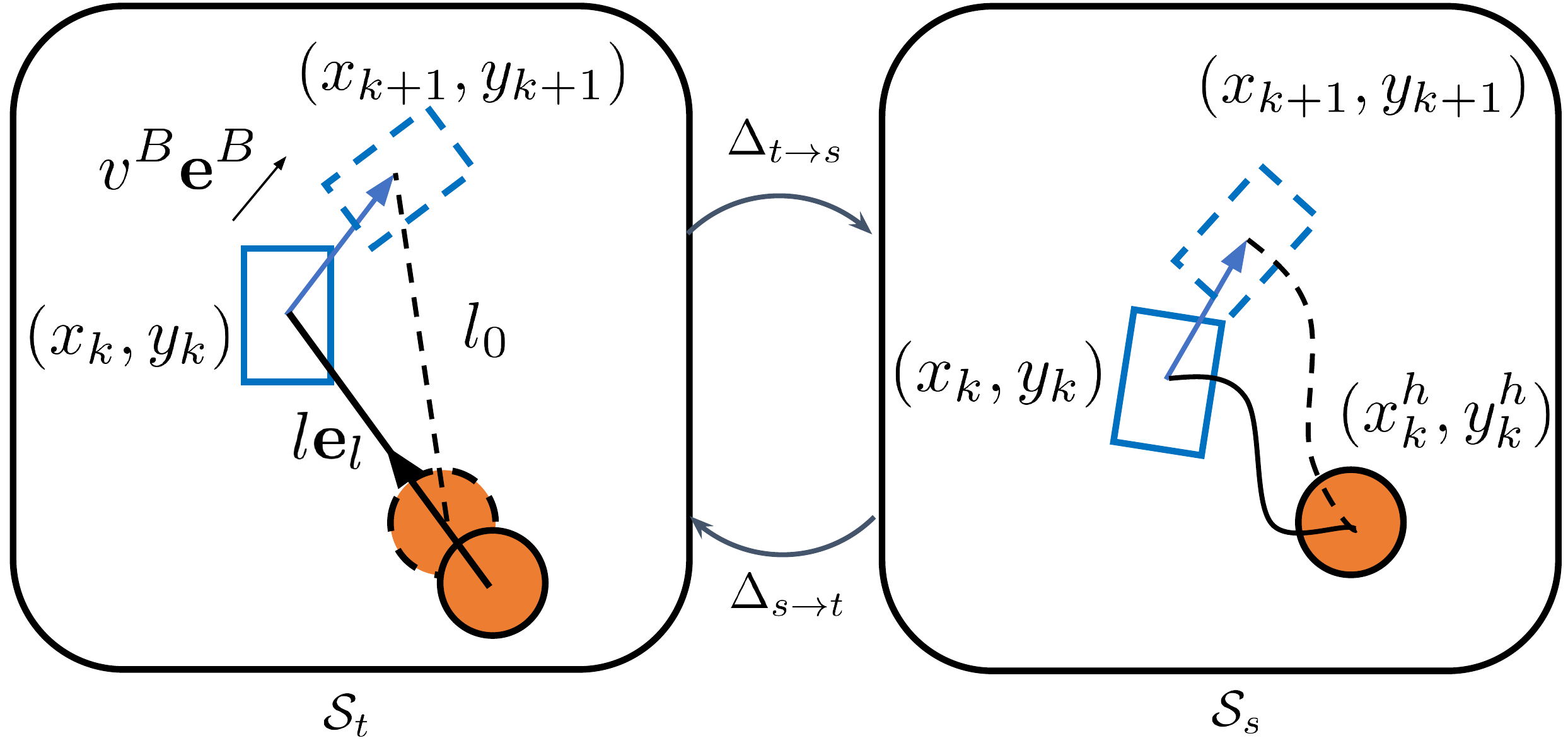}
    \caption{Hybrid modes switches in the discrete dynamics model $\Sigma$. The hybrid modes switches from $\mathcal{S}_s$ to $\mathcal{S}_t$ when leash becomes taut and the switch action is denoted as $\Delta_{s \rightarrow t}$. When the cable becomes slack, it switches back to $\mathcal{S}_s$ with action denoted as $\Delta_{t \rightarrow s}$. \update{This hybrid mode transitions are shown in the discrete-time manner.}}
    \vspace{-0.4cm}
    \label{fig:hybridsystem}
\end{figure}

When the leash is slack, the hybrid dynamics $\dot{\mathbf{q}}_s = f_{s}(\mathbf{q}_s,\mathbf{u}_s)$ can be defined as follows,
\begin{subequations}
\label{eq:leash-slack-dynamics}
\begin{align}
    \dot{x} &= v^B \mathbf{e}^B \cdot \mathbf{e}^W_x,\\
    \dot{y} &= v^B \mathbf{e}^B \cdot \mathbf{e}^W_y,\\
    \dot{\theta} &= \omega,\\
    \dot{\phi} &= - \| v^B \mathbf{e}_l \times \update{\mathbf{e}^B}\|/l_0, \label{eq:leash-slack-phi-dynamics}\\
    \dot{l} &= \update{v^B} \mathbf{e}_l \cdot \update{\mathbf{e}^B}, \label{eq:leash-slack-l-dynamics}
\end{align}
\end{subequations}
where \eqref{eq:leash-slack-phi-dynamics}, \eqref{eq:leash-slack-l-dynamics} comes from the geometric relation. where only the robot is moving since the cable is slack, shown in the right mode in Fig. \ref{fig:hybridsystem}.

\subsection{Leash Tension Model}
\label{subsec:leash-tension-model}
We seek a simple mapping from generalized coordinates to the leash tension, which allows for the consideration of physical interaction during the path planning.
To capture the relation between generalized coordinates and leash tension, we construct a linear regression model between robot speed and leash tension to minimize mean squared errors.
\begin{equation}
\label{eq:linear-regression-model}
    F = F_{\text{MSE}}(\update{\mathbf{q}_t}) = \beta_1 \mathbf{v}^B \cdot \mathbf{e}_l + \beta_2,
\end{equation}
where $\mathbf{v}^B \cdot \mathbf{e}_l$ represents the projected speed of robot along the taut leash direction and could be expressed by the generalized coordinates $\mathbf{q}_t$ with simple calculations, as proven by experiments in \ref{sec:experiments}.
%% ZL:11.1.2020 do we need say something about experiment? How do you know this is correct? It is a open question?

\subsection{Hybrid Modes Transition}
%% ZL:11.1.2020, say something about when the transition happens? and why you need model that. could be just single sentence. addressed
Since a taut leash is almost analogous to a rigid arm and thus infeasible in confined spaces due to the increased size of the human-robot system, the leash will need to transition to slack mode, and a hybrid system is introduced into the model.
We consider the following hybrid system $\Sigma$ as follows,
\begin{equation}
    \begin{split}
    \Sigma = \begin{cases}
                \dot{\mathbf{q}}_t = f_t(\mathbf{q}_t, \mathbf{u}_t), & \mathbf{q}_t \notin \mathcal{S}_s \\
                \mathbf{q}^{+}_s = \Delta_{s \rightarrow t}(\mathbf{q}_t^{-}), & \mathbf{q}_t^{-} \in \mathcal{S}_s \\
                \dot{\mathbf{q}}_s = f_s(\mathbf{q}_s, \mathbf{u}_s), & \mathbf{q}_s \notin \mathcal{S}_t \\
                \mathbf{q}^{+}_t = \Delta_{t \rightarrow s}(\mathbf{q}_s^{-}). & \mathbf{q}_s^{-} \in \mathcal{S}_t
             \end{cases}
    \end{split}
\end{equation}
The dynamics for two hybrid modes are shown in \eqref{eq:leash-taut-dynamics} and \eqref{eq:leash-slack-dynamics}. The two hybrid regions $\mathcal{S}_t$ and $\mathcal{S}_s$ are defined as follows,
\begin{align}
    \mathcal{S}_t &= \{(\update{\mathbf{q}_t}, F) \in \mathbb{R}^6: \mathbf{e}_{l} \cdot \mathbf{e}^{B} \geq 0 \wedge F \geq \bar{F} \} \\
    \update{\mathcal{S}_s} &= \{(\update{\mathbf{q}_s}, F) \in \mathbb{R}^6: \mathbf{e}_{l} \cdot \mathbf{e}^{B} \leq 0 \vee F \leq \bar{F} \}
\end{align}
where $\mathbf{e}_{l} \cdot \mathbf{e}^{B} < 0$, the robot and the human will approach each other in next time step which will make the leash slack. Moreover, $\bar{F}$ is applied as the lower bound representing the maximum intrinsic leash tension when the leash is slack.
\section{Path Planning with Hybrid Physical Human Robot Interaction}
\label{sec:local-planner}
%% ZL: 11.1.2020 Add one sentence about previous model
In this section, we discuss our optimization-based path planning algorithm where we consider the hybrid physical human robot interaction. We are given current coordinates $\mathbf{q}_{curr}$ and target goal coordinates $\mathbf{q}_{target}$ generated in the path from a global planner, which will be described in Sec. \ref{subsec:global-planner}.
A mixed-integer collocation-based problem along a horizon $N$ with time step $\Delta t = t/N$ is formulated as follows,

\begin{figure*}[t]
    \centering
    \includegraphics[width=0.85\linewidth]{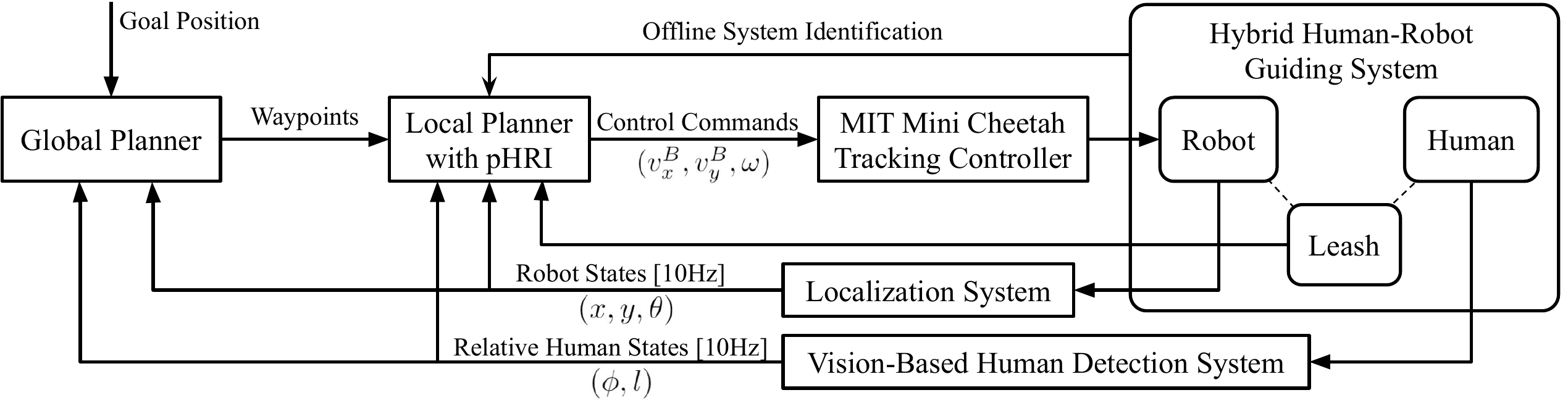}
    \caption{Framework for the Mini Cheetah robotic guide dog. Given the current position and goal position, a global planner using A* algorithm generates target feasible waypoints for the local planner. Then the local planner generates a short trajectory and sends control commands to the robot.}
    % \zli{need also mention astar and the entire data flow, whats the input/output of the each block}
    \label{fig:framework}
    \vspace{-0.3cm}
\end{figure*}

\begin{subequations}
\label{eq:direct-collocation}
\begin{align}
     \min\; & L(\mathbf{q}_k, \mathbf{u}_k, s_k, t) \quad \text{subject to} \label{eq:cost-function}\\ 
    & \mathbf{q}_{0} = \mathbf{q}_{curr}, \\
    & \mathbf{q}_{k+1} = \Bar{f}_t(\mathbf{q}_k, \mathbf{u}_k), \quad \text{if} \ s_k = 1 \label{eq:dynamics-constraint-taut} \\
    & \mathbf{q}_{k+1} = \Bar{f}_s(\mathbf{q}_k, \mathbf{u}_k), \quad \text{if} \ s_k = 0 \label{eq:dynamics-constraint-slack} \\
    & s_k = 1, \quad \text{if} \ \mathbf{e}_{l}(\mathbf{q}_k) \cdot \mathbf{e}^{B}(\mathbf{q}_k) \geq 0 \ 
    \wedge \ F_k \geq \bar{F}
    \label{eq:integer-constraint-taut} \\
    & s_k = 0, \quad \text{otherwise} \label{eq:integer-constraint-slack} \\
    & F_l(\mathbf{q}_k) \leq F_k \leq F_u(\mathbf{q}_k), \label{eq:force-constraint} \\
    & \mathbf{q}_l \leq \mathbf{q}_k \leq \mathbf{q}_u, \label{eq:state-constraint} \\
    & \mathbf{u}_l \leq \mathbf{u}_k \leq \mathbf{u}_u, \label{eq:input-constraint}
\end{align}
\end{subequations}

where $\Bar{f}_t$ and $\Bar{f}_s$ represent the discrete dynamics with sampling time step $\Delta t = t/N$. Furthermore, $s_k \in \{0, 1\}$ is the variable describing the hybrid dynamical mode in \eqref{eq:dynamics-constraint-taut} and \eqref{eq:dynamics-constraint-slack} and $s_k$ equals to one when the cable is taut and zero when the cable is slack, shown in \eqref{eq:integer-constraint-taut} and \eqref{eq:integer-constraint-slack}. The state, input and force constraints are imposed in \eqref{eq:state-constraint} and \eqref{eq:input-constraint} and \eqref{eq:force-constraint}. Notice that $\mathbf{e}_{l}$ and $\mathbf{e}^{B}$ could be expressed as functions of the generalized coordinates $\mathbf{q}_k$ at each time step, which was discussed in Sec. \ref{subsec:hybrid-dynamics}. 
%This nonlinear collocation-based optimization problem is solved with IPOPT \cite{biegler2009large} with CasADi \cite{andersson2019casadi} as the modelling language.
This nonlinear collocation-based optimization problem is formulated in CasADi~\cite{andersson2019casadi} with logic to switch modes and is solved with IPOPT \cite{biegler2009large}.

\subsection{Cost Function}
\label{subsec:cost-function}
We define the cost function \eqref{eq:cost-function} as follows,
\begin{equation}
    \begin{split}
        & L(\mathbf{q}_k, \mathbf{u}_k, s_k, t) = ||\mathbf{q}_N - \mathbf{q}_{target}||_{\mathbf{Q}_{target}} + S_t t \\
        & + \sum_{k = 0}^{N-1} \left(||\mathbf{u}_k||_{\mathbf{Q}_{\mathbf{u}}} + S_F F_k + S_l (l_0 - l_k) + S_{\Delta F} 
        (F_{k+1} - F_{k})\right)
    \end{split}
\end{equation}
where we have $\mathbf{Q}_{target} \in \mathbb{R}^5$, $\mathbf{Q}_{\mathbf{u}} \in \mathbb{R}^3$ as positive definite and $S_F, S_{\Delta F}, S_l$ as positive scalars. We have two terminal cost terms and four stage cost terms.
The term $||\mathbf{q}_N - \mathbf{q}_{target}||_{\mathbf{Q}_{target}}$ represents the quadratic terminal cost which tries to minimize the deviation of the final node from the target position. We do not assert a hard constraint for reaching the target position and it allows us more feasibility in the optimization. The term $S_t t$ allows us to find an optimal travel time for robot motions instead of using a fixed value.

For the stage cost, $||\mathbf{u}_k||_{\mathbf{Q}_{\mathbf{u}}}$ minimizes the control input, $S_F F_k$ minimizes the leash tension while ensuring smooth tension change with additional cost $S_{\Delta F} (F_{k+1} - F_{k})$. The term $S_l (l_0 - l_k)$ brings us faster optimization convergence and it tends to have more taut modes, which helps to guide the guided person since the person is immobile when $l_k < l_0$.

\subsection{Data-driven Leash Tension Constraint}
\label{subsec:force-constraint}
In our leash tension model \eqref{eq:linear-regression-model} in Sec. \ref{subsec:leash-tension-model}, we have seen that we have a mapping from the generalized coordinates to the leash tension. During the implementation of our collocation-based problem \eqref{eq:direct-collocation}, instead of adding this mapping relation as a constraint, we impose lower and upper bounds on this mapping, where we have
\begin{equation}
    F_l(\mathbf{q}_k), F_u(\mathbf{q}_k) = F_{\text{MSE}}(\mathbf{q}_k) \pm \sigma(F_{\text{MSE}}),
\end{equation}
where $\sigma(F_m)$ represents the standard deviation of the linear regression in our leash tension model \eqref{eq:linear-regression-model}.
Notice that imposing a two-sided constraint brings larger feasibility compared to an equality constraint.
This force constraint allows us to consider the physical human-robot interaction in the planner.

\subsection{Obstacle Avoidance}
\label{subsec:obstacle-avoidance}
When the system is required to navigate in an environment with obstacles, our optimization problem in \eqref{eq:direct-collocation} is no longer sufficient and obstacle avoidance constraints need to be added. 
In this paper, we consider the obstacle avoidance for both the robot and the human, where two simple signed distance constraints are imposed on them. Assume the $j$-th obstacle is located at $\mathbf{x}^{obs,j}_k = (x^{obs,j}_k, y^{obs,j}_k)$ at time step $k$. We then have,
\begin{align}
    ||\mathbf{x}_k - \mathbf{x}^{obs,j}_k || &\geq d + r + r^{obs}_j, \\
    ||\mathbf{x}^h_k - \mathbf{x}^{obs,j}_k || &\geq d + r^h + r^{obs}_j,
\end{align}
where $r$, $r^h$ and $r^{obs}_j$ represent the robot, human and obstacle dimensions. We also add a safety margin $d$ which allows us to ensure safety while handling our state-estimation and tracking errors.

% \begin{figure}
%     \centering
%     \includegraphics[width=1\linewidth]{img/state_switch.pdf}
%     \caption{The result of mode switching in planning.}
%     \label{fig:leash-tension-model}
% \end{figure}

\section{Quadrupedal Robotic Guide System}
\label{sec:framework}
% \begin{figure*}[htbp]
%     \centering
%     \includegraphics[width=0.95\linewidth]{img/framework.pdf}
%     \caption{Framework for the Mini Cheetah leading people.}
%     \label{fig:framework}
% \end{figure*}

\subsection{Framework}
\label{subsec:framework}
To safely navigate and guide a visually-blind person in a cluttered environment, an end-to-end framework is constructed and illustrated in Fig. \ref{fig:framework}. 
Our planner is composed of a search-based A* global planner and a collocation-based local planner with physical human robot interaction, as introduced in the previous section.
\subsection{Global Planner}
\label{subsec:global-planner}
For the global planner, we use a search-based A* planner over the grid map on a reduced generalized coordinates $\tilde{\mathbf{x}} = (x, y, \phi) \in \mathbb{R}^3$. 
The continuous transition between nodes on the $\mathbb{R}^3$ configuration space is defined as $(\Delta x, \Delta y, \Delta \phi)$. For experiments, we have $\Delta x = \pm 0.5$, $\Delta y = \pm 0.25$ and $\Delta \phi = \pm \pi/{8}$. The node cost and heuristic  cost to-go at node $\tilde{\mathbf{x}}_n$ are defined as $g(\tilde{\mathbf{x}}_n)$ and $h(\tilde{\mathbf{x}}_n)$ where, 
\begin{equation}
    g(\tilde{\mathbf{x}}_n) = \sum_{i = 1}^{N-1} ||\tilde{\mathbf{x}}_n - \tilde{\mathbf{x}}_{n-1}||^2
\end{equation}
\begin{equation}
    \begin{split}
        h(\tilde{\mathbf{x}}_n) =& ||x_N - x_{goal}||^2 + ||y_N - y_{goal}||^2 +  \\
        & ||\phi_N - \phi_{goal}||^2 + \lambda (1 - \cos(\theta_N - \theta_{goal})), 
    \end{split}
\end{equation}
where $\theta_N$ can be calculated with approximate dynamics \eqref{eq:dynamics-constraint-taut} using continuous transition between nodes while assuming the leash is always taut. A cosine function is applied on $\theta$ in the heuristic cost to-go to solve the singularity problem.
This A* global planner generates a continuous collision-free trajectory with a sequence of 5-dimensional waypoints. 
This is passed to the local planner with pHRI introduced in Sec.~\ref{sec:local-planner}.
% Based on the hybrid dynamical model illustrated in Sec. \ref{sec:hybrid-pHRI-model}, our local optimization-based human-robot interactive planner using the target way points and the estimated human-robot current state to plan next 2-4 seconds' trajectory, which needs to consider the kinematic feasibility as well as the obstacle avoidance, illustrated in Sec. \ref{sec:local-planner}.
% The Mini Cheetah Robot executed the optimized commands $(\mathbf{v}^B, \omega)$ when following the trajectory above and execute zero commands when that robot reaches near the terminal point of the trajectory, which allows the robot to ensure the safety even with a time-delay from the local planner.

% The localization system can get the current robot state $(x, y, \theta)$ and the vision-based human detection system can get the relative polar coordinates $l$ and $\phi$ in robot frame. With geometric relationship between the robot and the human, we can calculate our generalized coordinates $\mathbf{q}$ in the world frame.

\subsection{Robot Localization And Human Tracking}
\label{subsec:localization-tracking}
% In this section, we show our robot localization and human tracking method. 
Knowing robot and human states online is critical for the autonomous system.
We firstly build an occupancy grid map with a 2D lidar based on Hector Slam \cite{hector}. 
Later, AMCL\cite{amcl} is utilized to estimate the robot states in the world frame.
To estimate the position of the guided person, a Depth-RGB camera is used to detect the human's 3D position through OpenVINO \cite{Gorbachev_2019_ICCV} and a Kalman Filter based on a constant-speed linearized system \cite{zhongyu_2019} is applied to track the detected human position.
The camera is deployed on a 2 DoF gimbal which can rotate and pitch. This camera gimbal is mounted on the robot's top surface and is programmed to keep the guided person visible in the camera frame irrespective of the relative orientation of the human with respect to the robot.

\subsection{Velocity Tracking Controller on Mini Cheetah}
We use the existing state of the art velocity tracking controller for Mini Cheetah, where a MPC \cite{di2018dynamic} computes desired ground reaction forces and desired foot and body position from given velocity commands. From these desired contact forces, WBIC \cite{kim2019highly} computes desired joint position velocity that are delivered to joint-level controllers to generate joint torques.

% \subsection{System Identification}
% \label{subsec:system-identification}
% The hybrid kinematic models and leash tension model are estimated with an offline system identification from the random trails from the proposed human-leash-robot system, where we find the discount coefficients $\alpha_x$, $\alpha_y$, $\alpha_{\theta}$, $\alpha_{\phi}$~\eqref{eq:leash-taut-dynamics} in Sec.~\ref{subsec:hybrid-kinematics} and use a linear regression to get the $\beta_1$ and $\beta_2$~\eqref{eq:linear-regression-model} in Sec.~\ref{subsec:leash-tension-model} in order to have best estimation and prediction over the physical human-robot interaction.
\section{Experiments and Evaluation}
\label{sec:experiments}
% \subsection{Hardware Setup}
% Our experiments were conducted on Mini Cheetah both indoor and outdoor \jun{(TBD)}. As shown in Fig. \ref{}, we using a slack leash connect robot and human, the robot has several sensors onboard - including a 100N range force sensor (load cell) on the leash to measure interactive force between robot and human; an Intel RealSense D435i depth camera with 2-axis (pan and tilt) servo gimbal mounted on the top tracking the human position relevant to the robot; a SLAMTEC 2D rplidar-A1 for global localization and mapping; an amazon Echo dot for speech interaction. 

% An Arduino used for transferring force data from force sensor (load cell) to NUC, and also receiving command from NUC then actuating servo gimbal.
%% ZL:11.1.2020 Is this necessary? addressed
The hardware setup is illustrated in Fig. \ref{fig:cover}. and all the aforementioned proposed algorithms are running on an onboard Intel computer using ROS, while the velocity tracking controller is running on a real-time computer within the Mini Cheetah. 
    
\subsection{Offline System Identification}
\subsubsection{Human-robot dynamic model}

\begin{figure}%[htbp]
    \begin{subfigure}[t]{0.95\linewidth}
        \centering
        \includegraphics[width=0.48\linewidth]{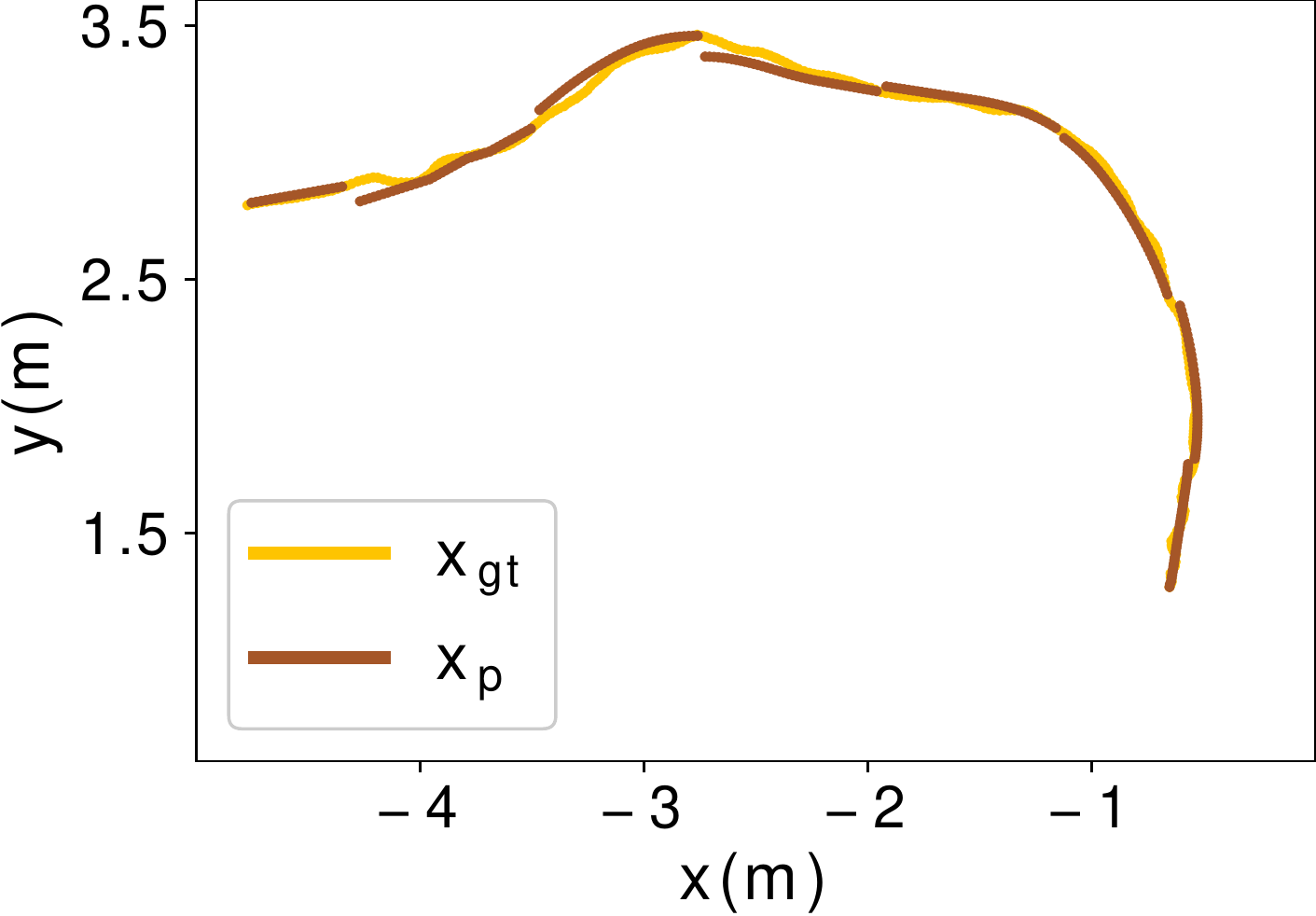}
        \includegraphics[width=0.48\linewidth]{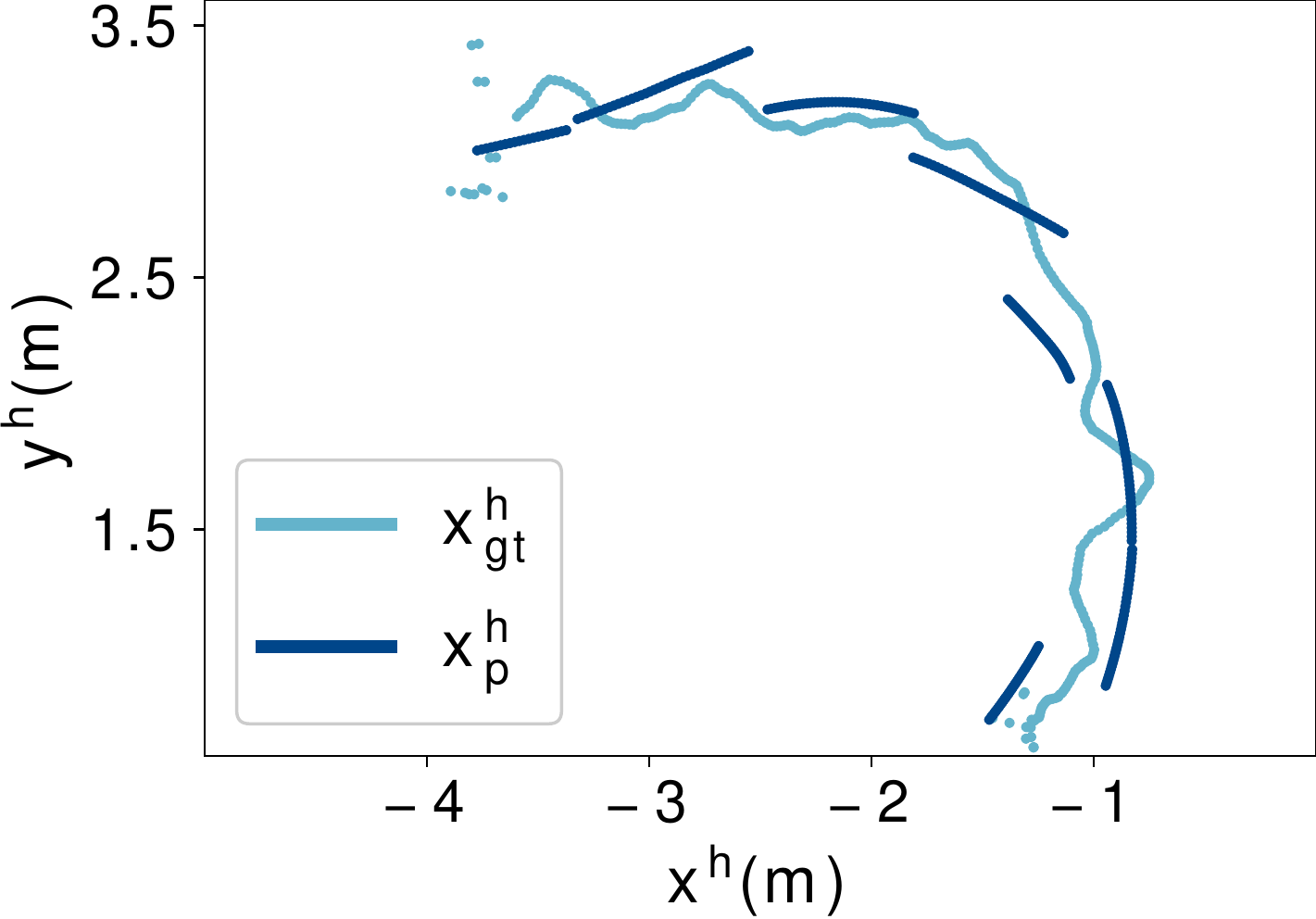}
        \vspace{-0.2cm}
        \caption{}
        \label{subfig:alpha-demo-1}
    \end{subfigure}\\
    \begin{subfigure}[t]{0.95\linewidth}
        \centering
        \includegraphics[width=0.48\linewidth]{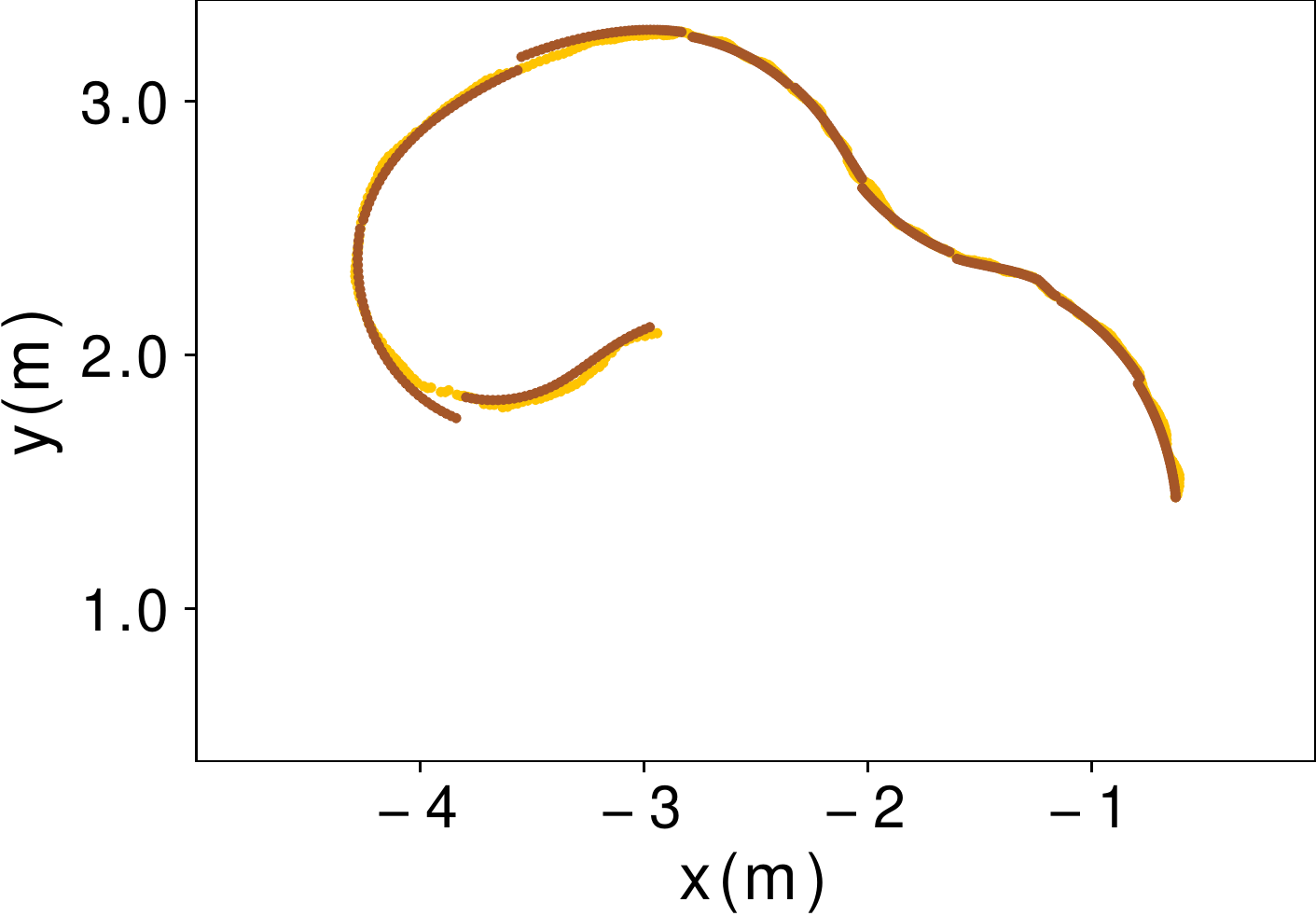}
        \includegraphics[width=0.48\linewidth]{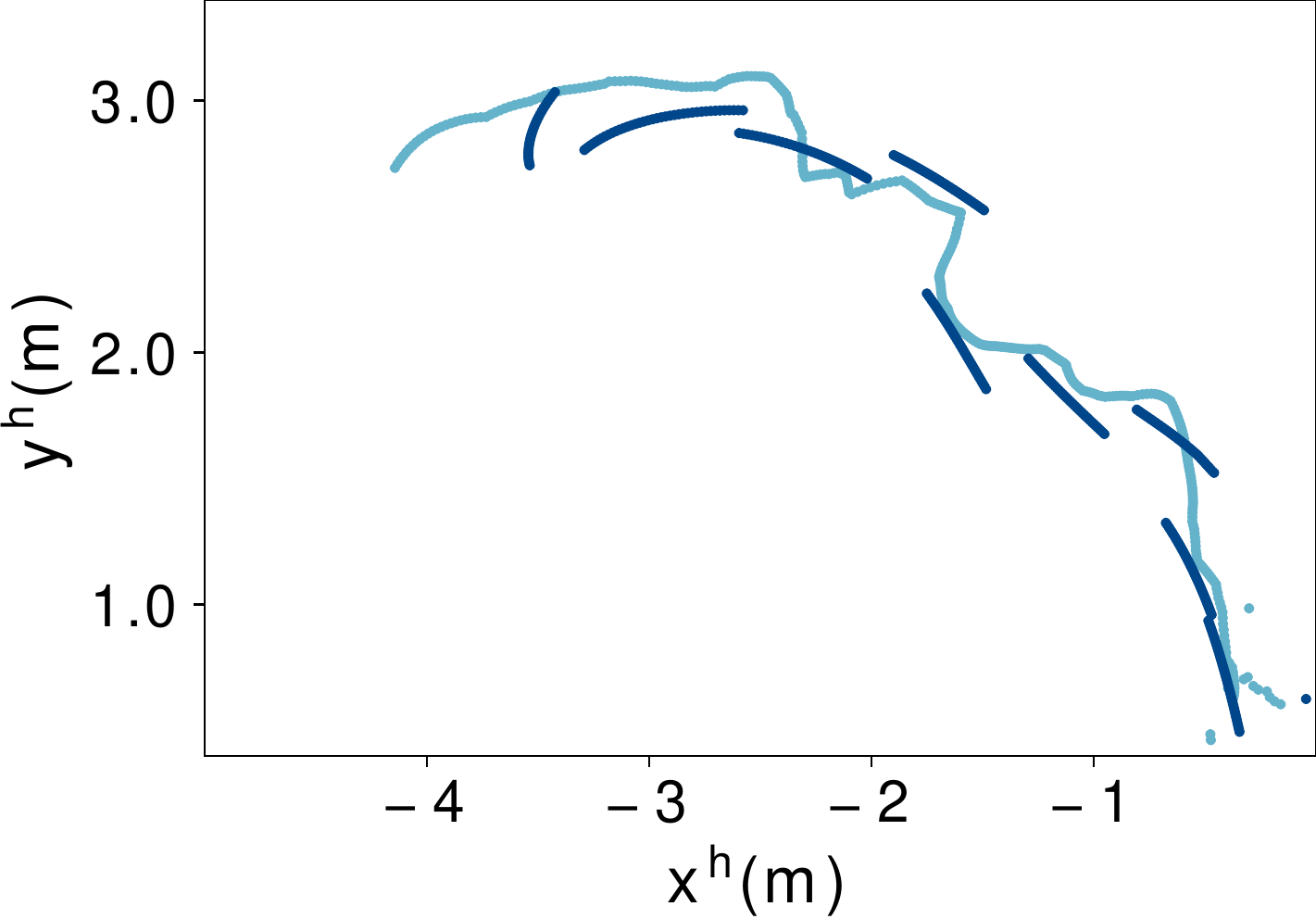}
        \vspace{-0.2cm}
        \caption{}
        \label{subfig:alpha-demo-2}
    \end{subfigure}
    \caption{Two typical guiding test cases for optimizing the human-robot dynamic model by minimizing the error of ground truth robot position $\mathbf{x}_{gt}$ and model-based computed robot position $\mathbf{x}_p$. Fig. \ref{subfig:alpha-demo-1} and Fig. \ref{subfig:alpha-demo-2} are two separate test cases, with the robot trajectory on the left and the human trajectory on the right.}
    \vspace{-0.3cm}
    \label{fig:trajectory-benchmark}
\end{figure}

The discount coefficients $\bm{\alpha}$ in \eqref{eq:alpha-vector} in Sec.~\ref{subsec:hybrid-dynamics} for the taut mode of human-robot dynamic are first identified based on offline experimental data. In order to obtain such data, several blind-folded people are guided in an unknown environment along various trails. 
In each trail, a human operator commands the Mini Cheetah to randomly move around while ensuring a taut leash.
In such experiments, robot global positions $\mathbf{x}_{gt}$ and human positions $\mathbf{x}^h_{gt}$ are recorded along the sample trajectories, serving as ground truth data, and $\bm{\alpha}$ is sampled in the range of $[0, 1]$.
With each set of these sampled values, we compute the predicted robot global positions, denoted as $\mathbf{x}_{p}$, based on the dynamic model of the taut mode \eqref{eq:leash-taut-dynamics}.
The value of $\bm{\alpha}$ that can produce the smallest least mean squared distance between the predicted and estimated trajectories are picked.
The identified value of $\bm{\alpha}$ is $[0.8, 0.8, 0.6, 0.8]$, which was obtained by minimizing the prediction error for the human-robot system. This is incorporated in the dynamic model for later human guiding experiments.

The ground truth and predicted trajectories are illustrated by the identified $\bm{\alpha}$ in Fig. \ref{fig:trajectory-benchmark}. The identification of the dynamic model of taut mode matches well between the ground truth robot position $\mathbf{x}_{gt}$ and the predicted robot position $\mathbf{x}_p$, with an average prediction error of 0.023m. The human prediction error is 0.176m, which is acceptable, considering the noise of human detection and estimation. 
%\zli{why is tracking error? why not prediction error?}
% \zli{what about human? trajectories describe coordinates, shall we say that the generalized coordinates match well?}

% Moreover, as shown in the last part of Fig.~\ref{fig:framework}, a constant value for $\bm{\alpha}$ may not work ideally in all scenarios. Therefore, we also take the maximum prediction error due to this model mismatch into the safety margin in the proposed optimization-based local planner. \zli{what is the maximum prediction erro? do mention it}
% This allows to minimize the model mismatch while considering the human-robot interaction in our system while ensuring the safety in our optimization-based local planner.
%% ZL:11.1.2020 NEED average prediction error value! why pick 0.8?  @tong wenzhe

\subsubsection{Leash Tension Model}

\begin{figure}
    \centering
    \includegraphics[width=0.95\linewidth]{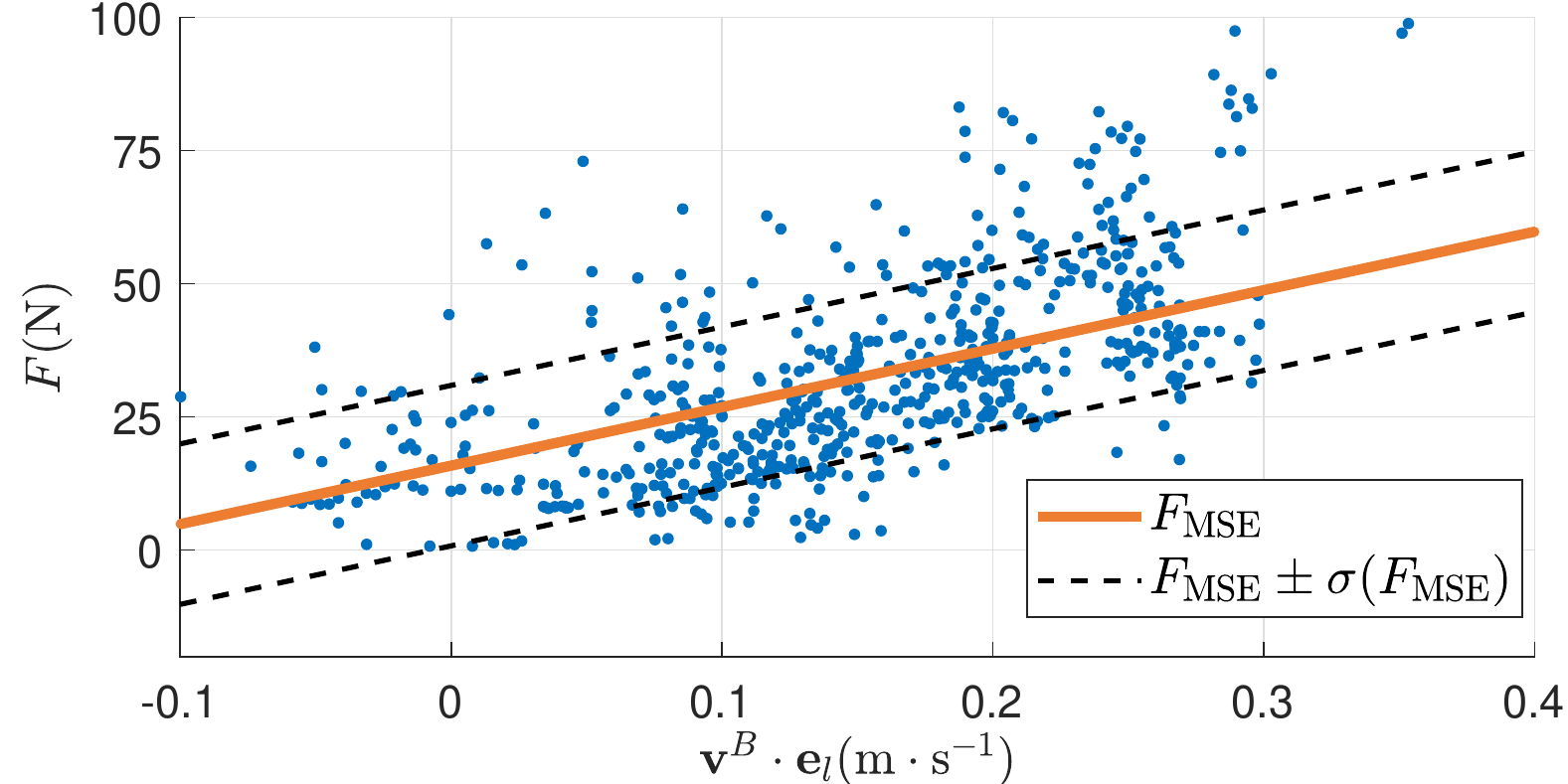}
    \caption{Validating leash tension model in Eq.~\eqref{eq:linear-regression-model} by randomly walking. Linear regression is carried out between the projected speed in the leash direction $\mathbf{v}^B \cdot \mathbf{e}_l$ and leash tension $F$. As shown, 76\% of the measured tension data lies between $F_m \pm \sigma(F_m)$. Furthermore, $\sigma(F_m)$ represents the standard deviation of this linear regression.}
    \label{fig:leash-tension-model}
    \vspace{-0.3cm}
\end{figure}

The leash tension model given by \eqref{eq:linear-regression-model} in Sec.~\ref{subsec:leash-tension-model} is validated by letting the robot guide a human via a leash to move randomly, with the leash being either slack and taut. The interactive force $F$, system states and control commands $(\mathbf{v}^B, \phi)$ are recorded. The projected velocity along the leash direction $\mathbf{v}^B \cdot \mathbf{e}_l(\phi)$ is later obtained. 

\begin{figure*}[tb]\label{fig: experiemnt results}
    \centering
    \begin{subfigure}[t]{0.24\linewidth}
        \centering
        \includegraphics[height = 4.6cm]{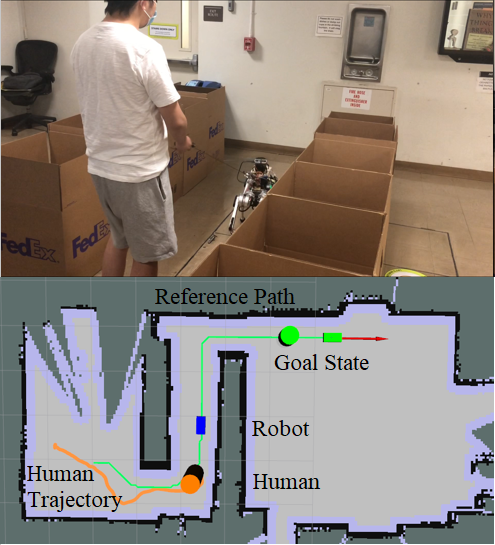}
        \caption{}\label{subfig:straight}
        % (a)
    \end{subfigure}
    \begin{subfigure}[t]{0.24\linewidth}
        \centering
        \includegraphics[height = 4.6cm]{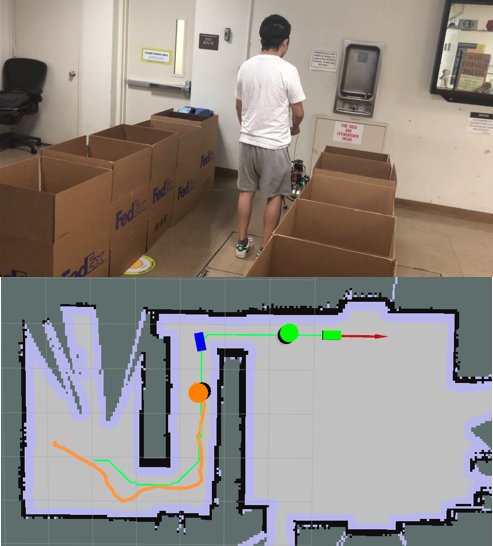}
        \caption{}\label{subfig:begin-coner}
        % (b)
    \end{subfigure} 
    \centering
    \begin{subfigure}[t]{0.24\linewidth}
        \centering
        \includegraphics[height = 4.6cm]{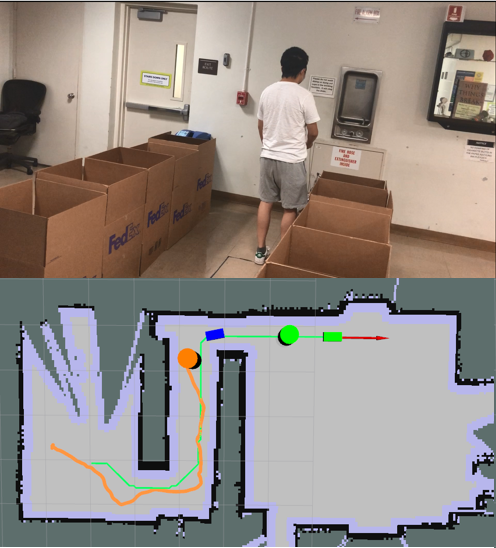}
        \caption{}\label{subfig:in-coner}
        % (c)
    \end{subfigure}
    \begin{subfigure}[t]{0.24\linewidth}
        \centering
        \includegraphics[height = 4.6cm]{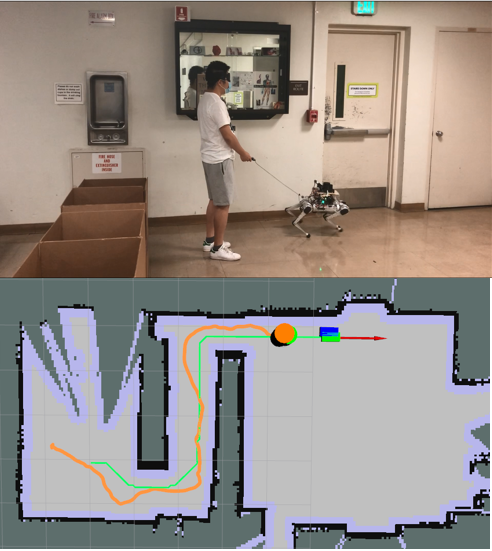}
        \caption{}\label{subfig:pass-sconer}
        % (d)
    \end{subfigure}
    \caption{Snapshot of the guiding process. The Mini Cheetah guides a blindfolded person from an initial to a target position. The blue marker and orange marker represent the robot and human. Human history trajectory is represented by the orange line while the robot global planned reference is represented by the green line. }
    % (a): Human is guided by the robot with taut leash in the straight corridor. (b): The system approaches the narrow space, and the robot guide the human to a feasible place. (c): The leash becomes slack and robot changes its position and direction without pulling the human. (d): The human-robot system successfully switches back to taut mode and passes the narrow doorway.}
    
    \label{fig:experiment-snapshots}
    \vspace{-0.3cm}
\end{figure*}

We fit our force $F$ with a linear model \eqref{eq:linear-regression-model} while minimizing the least square errors, shown in Fig. \ref{fig:leash-tension-model}. The reasons for choosing a linear model over higher-order ones are two-fold: during experiments higher-order models did not exhibit superior prediction performance and a linear model reduces complexity in the optimization-based local planner.  The parameters we optimized are $\beta_1 = 109.8$ and $\beta_2 = 15.85$. The standard deviation of force is $\sigma(F_{\text{MSE}}) = 15.06$, which was used for estimating lower and upper bounds of force constraints \eqref{eq:force-constraint}. We verify that 76\% of our force data lies between the region constrained with our linear regression model and related standard deviation, which is acceptable for estimating the force constraints.

\subsection{Robot Guiding Human Experiments}
The proposed system is evaluated in various maps of narrow spaces.
One such example is shown in Fig. \ref{fig:experiment-snapshots}, where the map consists of a narrow doorway connected to a narrow corridor, with the narrowest traversal region being only of width 1.0~m. Since the human-robot system has a length of 1.6 m when the leash is taut and the human only moves along the direction of the force, it is hard for the human-robot system to pass through this region if the leash stays taut. This allows for the demonstration of the hybrid mode switch in our local planner. This map contains several situations the system will face in the real world including doors, narrow corridors and corners.

The experimental goal is to enable the Mini-cheetah to safely guide a blindfolded person to the given goal location $\mathbf{q}_{goal}$ without colliding with obstacles. To evaluate the performance of our planning system, we choose the several different goal locations far from the different initial locations and let the robot plan and control fully autonomously. Three adults participated in the experiments in this narrow map.\\
% Our global planner get the goal position choose by the human and plan a global human-robot path. The local planner plan the local trajectory of 2-3 s with the refresh time of 3s. The robot command generated by local planner is transposed to the robot with 100hz.
In the experiments, the human-robot system successfully reached the given goal without any collision.

For the example experiment shown in Fig. \ref{fig:experiment-snapshots}, the time the Mini Cheetah took to guide the blindfolded person to the random final goal position is roughly 75s. In this map, the leash switched to taut at the beginning part of the task. When the human-robot system came to the most narrow region of the second doorway, the leash switched to slack mode and the guided human stopped moving as shown in Fig. \ref{subfig:in-coner}. After the robot changed its configuration that allowed it to guide the human pass the narrow region, the human-robot system switched to the taut mode and passed this doorway as shown in Fig. \ref{subfig:pass-sconer}.

\begin{figure}[htbp]
    \begin{subfigure}[t]{0.48\linewidth}
        \centering
        \includegraphics[width=\linewidth]{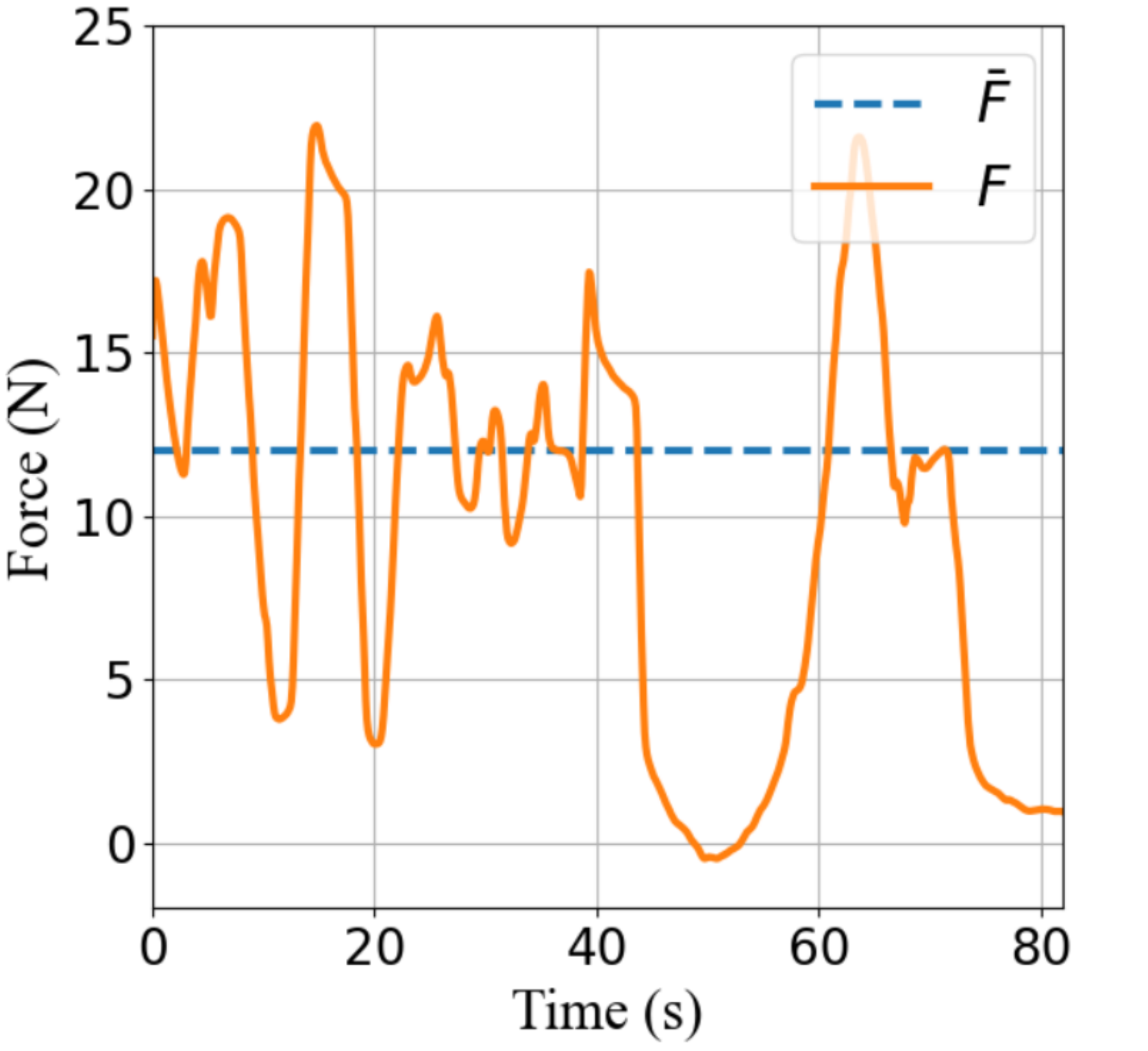}
        \caption{The tension in the leash}
        \label{subfig:experiment-leash-tension}
    \end{subfigure} 
    \begin{subfigure}[t]{0.48\linewidth}
        \centering
        \includegraphics[width=\linewidth]{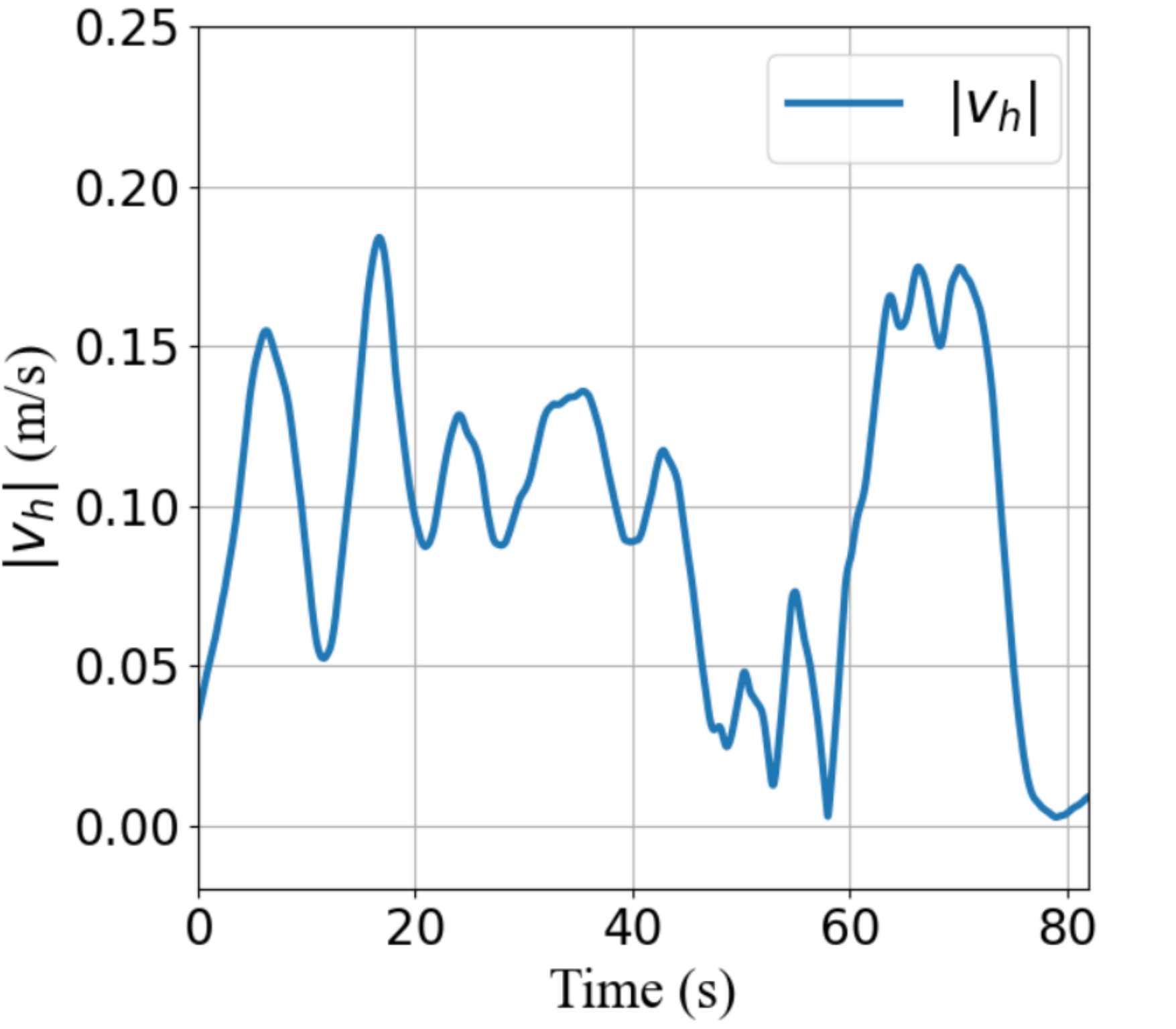}
        \caption{Speed of Human}
        \label{subfig:experiment-human-speed}
    \end{subfigure}
    \label{fig:experiment}
    \caption{Experimental results of the tension in the leash and human speed. $\bar{F}$ represents the threshold of the force. From 45s-60s, the force in the leash in (a) is nearly 0 N (implicating a slack leash) and the human moving speed is also below 0.05 m/s in (b).}
    \vspace{-0.3cm}
\end{figure}

Moreover, as shown in Fig. \ref{subfig:experiment-leash-tension}, the tension threshold $\bar{F}$ measured at the beginning of experiment is 12 N. We notice that from 45s  to 60s, when the human-robot system approached the narrow space of the second doorway, the force in the leash was extremely small as the system switched into slack mode, and the robot was changing its individual configuration until it was able to guide the human pass the doorway. In this period, the human was not pulled by the robot and stopped moving, as shown by the fact that the speed of human movement was near zero (0.05m/s) between 45s to 60s. After 60s, the robot changed its position and orientation to a suitable state, switching to taut mode to apply the leash force again to guide the human to the final goal position. 

%Moreover, when compared with other time periods, the relationship between the change of force and the behaviour change of human movement is clear. From 15~s to 20~s, when the human-robot system tried to pass the first wide doorway, the force dropped below  $\bar{F}$ for a short moment, which caused the human to slow down. 

% \subsubsection{Hybrid Modes Switch with Interactive Force}
% One typical experiment result was chosen to demonstrate the human-robot hybrid mode switch with interactive force. The $\bar{F}$ calibrated in the begin of experiment was 12 N. The force measurement was recorded with a mean filter during the whole task. Also, to verify the relationship between human moving behavior and force applied to the human, we calculated the speed of human in the global frame as show in \ref{subfig:experiment-human-speed}.

% \begin{figure}[htbp]
%     \begin{subfigure}[t]{0.995\linewidth}
%         \centering
%         \includegraphics[width=0.995\linewidth]{img/force.png}
%         \caption{The tension in the leash}
%         \label{subfig:experiment-leash-tension}
%     \end{subfigure} \\
%     \begin{subfigure}[t]{1\linewidth}
%         \centering
%         \includegraphics[width=1\linewidth]{img/v_norm.PNG}
%         \caption{The human moving speed }
%         \label{subfig:experiment-human-speed}
%     \end{subfigure}
%     \label{fig:experiment}
%     \caption{Experimental result of the tension in the leash and human moving speed. From 45s-60s, the force in the leash in (a) is nearly 0 m/s and the human moving speed also below 0.05 m/s in (b).}
% \end{figure}
\section{Conclusion And Future Work}
To our knowledge, this work proposes one of the first end-to-end human-robot interaction system to serve as an autonomous navigation aid to enable the visually impaired to traverse narrow and cluttered spaces. A 
%for the hybrid physical human navigation aids based on the 
data-driven interaction force model and a hybrid dynamic model were developed to help plan paths with hybrid mode switches to switch between taut and slack states for the leash. 
%Based on the taut and slack state of the leash, 
A global planner along with a mixed-integer optimization-based local planer were formulated to generate trajectories that served as input to the low-level controller on the Mini Cheetah.
%The local planning task was formulated as a mixed-integer optimization problem. 
The proposed pipeline was deployed on the Mini Cheetah and validated by experiments with a blind-folded person. Experimental results indicate that our system is able to physically guide the person with a safe and efficient trajectory in a narrow space, including obstacle avoidance maneuvers and hybrid state transitions. Future work will focus on more complicated modeling of human behavior with force traction to propose more novel applications of the robotic guide dog.

%% ACKNOWLEDGEMENTS
\section*{Acknowledgement}
This work is supported in part by the National Science Foundation Grants CMMI-1944722. The authors would also like to thank Professor Sangbae Kim, the MIT Biomimetic Robotics Lab\update{, and NAVER LABS} for providing the Mini Cheetah simulation software and lending the Mini Cheetah for experiments.

\balance
%% BIBLIOGRAPHY
{
\bibliographystyle{IEEEtran}
\bibliography{IEEEabrv, bib/bibliography}
\balance
}

\end{document}